\documentclass[conference]{IEEEtran}
\IEEEoverridecommandlockouts
\usepackage{cite}
\usepackage{amsmath,amssymb,amsfonts}
\usepackage{algorithmic}
\usepackage{graphicx}
\usepackage{textcomp}
\usepackage{url}
\usepackage{xcolor}
\def\BibTeX{{\rm B\kern-.05em{\sc i\kern-.025em b}\kern-.08em
    T\kern-.1667em\lower.7ex\hbox{E}\kern-.125emX}}

\makeatletter
\def\ps@IEEEtitlepagestyle{
  \def\@oddfoot{\mycopyrightnotice}
  \def\@evenfoot{}
}
\def\mycopyrightnotice{
\begin{minipage}{\textwidth}
{\footnotesize
    \textcopyright 2019 IEEE. Personal use of this material is permitted. Permission from IEEE must be obtained for all other uses, in any current or future media, including reprinting/republishing this material for advertising or promotional purposes, creating new collective works, for resale or redistribution to servers or lists, or reuse of any copyrighted component of this work in other works.
\hfill}
\end{minipage}
  \gdef\mycopyrightnotice{}
}

\begin{document}

\title{Real-world Conversational AI for Hotel Bookings}

\author{\IEEEauthorblockN{Bai Li}
\IEEEauthorblockA{\textit{University of Toronto}\\
\textit{SnapTravel}\\
Toronto, Canada \\
bai@cs.toronto.edu}
\and
\IEEEauthorblockN{Nanyi Jiang}
\IEEEauthorblockA{\textit{SnapTravel}\\
Toronto, Canada \\
leon@snaptravel.com}
\and
\IEEEauthorblockN{Joey Sham}
\IEEEauthorblockA{\textit{SnapTravel}\\
Toronto, Canada \\
joey@snaptravel.com}
\and
\IEEEauthorblockN{Henry Shi}
\IEEEauthorblockA{\textit{SnapTravel}\\
Toronto, Canada \\
henry@snaptravel.com}
\and
\IEEEauthorblockN{Hussein Fazal}
\IEEEauthorblockA{\textit{SnapTravel}\\
Toronto, Canada \\
hussein@snaptravel.com}
}

\maketitle

\begin{abstract}
In this paper, we present a real-world conversational AI system to search for and book hotels through text messaging.
Our architecture consists of a frame-based dialogue management system, which calls machine learning models for intent classification, named entity recognition, and information retrieval subtasks.
Our chatbot has been deployed on a commercial scale, handling tens of thousands of hotel searches every day.
We describe the various opportunities and challenges of developing a chatbot in the travel industry.
\end{abstract}

\begin{IEEEkeywords}
conversational AI, task-oriented chatbot, named entity recognition, information retrieval
\end{IEEEkeywords}

\section{Introduction}

\begin{figure}[t]
    \centering
    \includegraphics[width=0.65\linewidth]{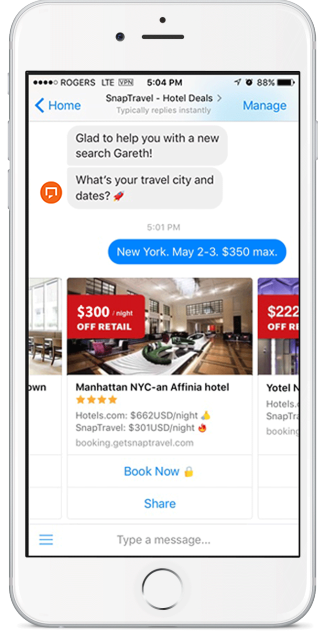}
    \caption{Screenshot of a typical conversation with our bot in Facebook Messenger.}
    \label{fig:screenshot_phone}
\end{figure}

Task-oriented chatbots have recently been applied to many areas in e-commerce. In this paper, we describe a task-oriented chatbot system that provides hotel recommendations and deals. Users access the chatbot through third-party messaging platforms, such as Facebook Messenger (Figure \ref{fig:screenshot_phone}), Amazon Alexa, and WhatsApp. The chatbot elicits information, such as travel dates and hotel preferences, through a conversation, then recommends a set of suitable hotels that the user can then book. Our system uses a dialogue manager that integrates a combination of NLP models to handle the most frequent scenarios, and defer to a human support agent for more difficult situations.

The travel industry is an excellent target for e-commerce chatbots for several reasons:

\begin{enumerate}
    \item Typical online travel agencies provide a web interface (such as buttons, dropdowns, and checkboxes) to enter information and filter search results; this can be difficult to navigate. In contrast, chatbot have a much gentler learning curve, since users interact with the bot using natural language. Additionally, chatbots are lightweight as they are embedded in an instant messaging platform that handles authentication. All of these factors contribute to higher user convenience \cite{rise-of-bots}.
    \item Many people book vacations using travel agents, so the idea of booking travel through conversation is already familiar. Thus, we emulate the role of a travel agent, who talks to the customer while performing searches on various supplier databases on his behalf.
    \item Our chatbot has the advantage of a narrow focus, so that every conversation is related to booking a hotel. This constrains conversations to a limited set of situations, thus allowing us to develop specialized models to handle hotel-related queries with very high accuracy.
\end{enumerate}

\begin{figure}
    \centering
    \includegraphics[width=0.9\linewidth]{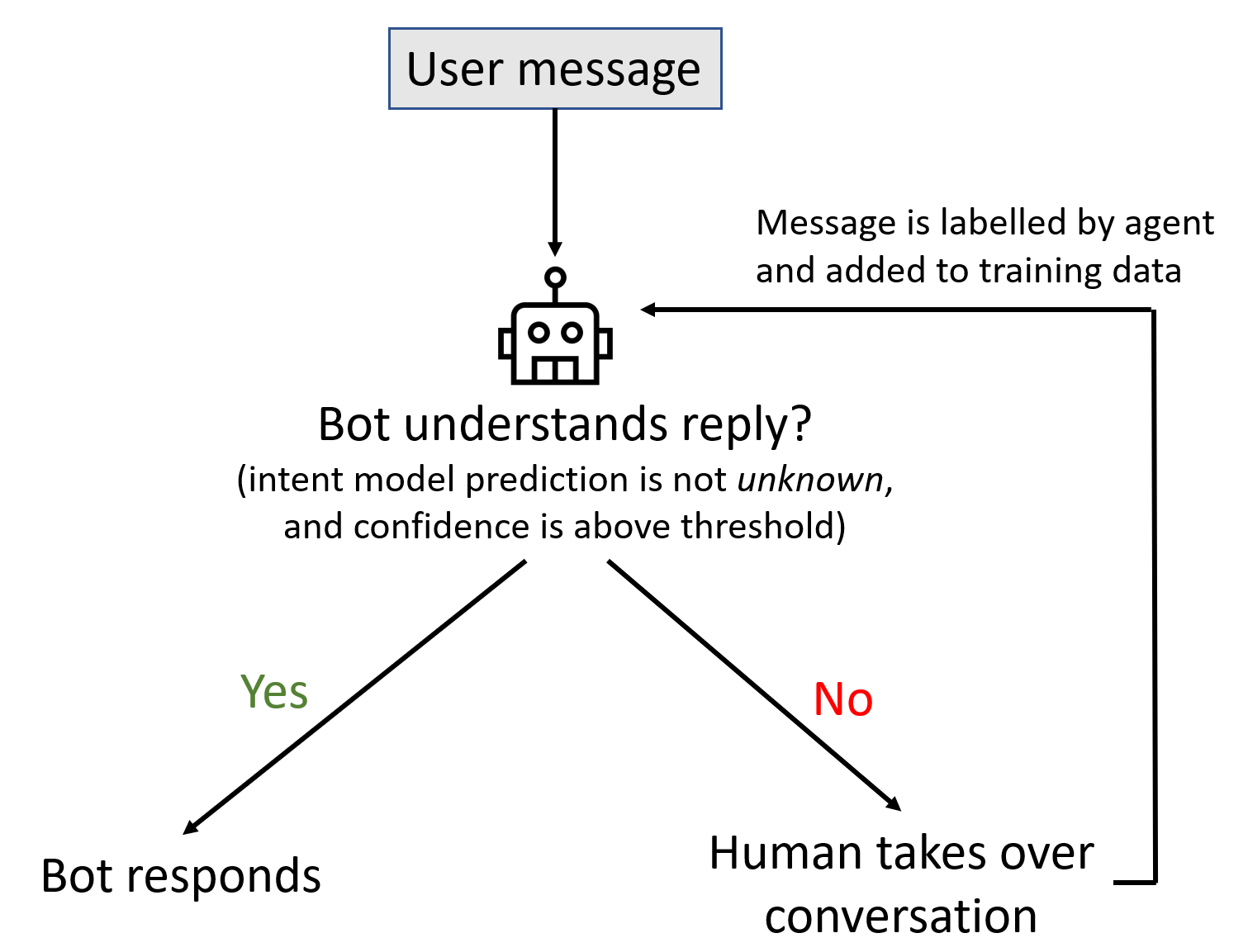}
    \caption{The intent model determines for each incoming message, whether the bot can respond adequately. If the message cannot be recognized as one of our intent classes, then the conversation is handed to a human agent, and is added to our training data.}
    \label{fig:bot-understands}
\end{figure}

The automated component of the chatbot is also closely integrated with human support agents: when the NLP system is unable to understand a customer's intentions, customer support agents are notified and take over the conversation. The agents' feedback is then used to improve the AI, providing valuable training data (Figure \ref{fig:bot-understands}). In this paper, we describe our conversational AI systems, datasets, and models.

\section{Related work}

Numerous task-oriented chatbots have been developed for commercial and recreational purposes. Most commercial chatbots today use a frame-based dialogue system, which was first proposed in 1977 for a flight booking task \cite{gus-system}. Such a system uses a finite-state automaton to direct the conversation, which fills a set of slots with user-given values before an action can be taken. Modern frame-based systems often use machine learning for the slot-filling subtask \cite{slot-filling-lstm}.


Natural language processing has been applied to other problems in the travel industry, for example, text mining hotel information from user reviews for a recommendation system \cite{zhang-hotel-recsys}, or determining the economic importance of various hotel characteristics \cite{ghose-ranking}. Sentiment analysis techniques have been applied to hotel reviews for classifying polarity \cite{shi-sentiment} and identifying common complaints to report to hotel management \cite{kasper-sentiment}.

\section{Chatbot architecture}

Our chatbot system tries to find a desirable hotel for the user, through an interactive dialogue. First, the bot asks a series of questions, such as the dates of travel, the destination city, and a budget range. After the necessary information has been collected, the bot performs a search and sends a list of matching hotels, sorted based on the users' preferences; if the user is satisfied with the results, he can complete the booking within the chat client. Otherwise, the user may continue talking to the bot to further narrow down his search criteria.

At any point in the conversation, the user may request to talk to a customer support agent by clicking an ``agent'' or ``help'' button. The bot also sends the conversation to an agent if the user says something that the bot does not understand. Thus, the bot handles the most common use cases, while humans handle a long tail of specialized and less common requests.

The hotel search is backed by a database of approximately 100,000 cities and 300,000 hotels, populated using data from our partners. Each database entry contains the name of the city or hotel, geographic information (e.g., address, state, country), and various metadata (e.g., review score, number of bookings).

\subsection{Dialogue management}

\begin{figure*}
    \centering
    \includegraphics[width=0.8\linewidth]{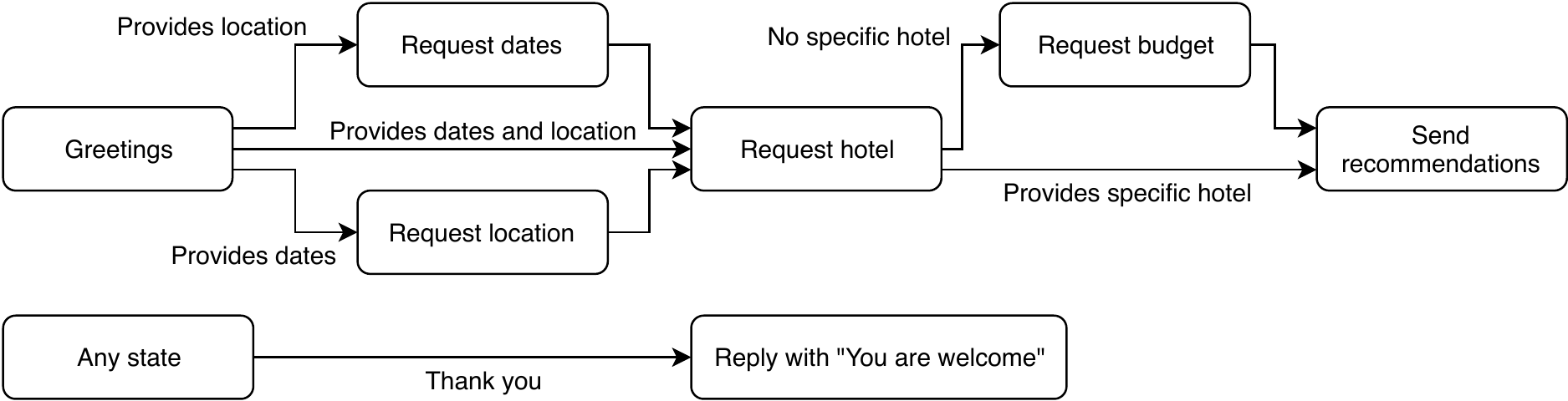}
    \caption{Diagram showing part of the state machine, with relevant transitions; this part is invoked when a user starts a new search for a hotel.}
    \label{fig:state-machine}
\end{figure*}

\begin{figure*}
    \centering
    \includegraphics[width=0.8\linewidth]{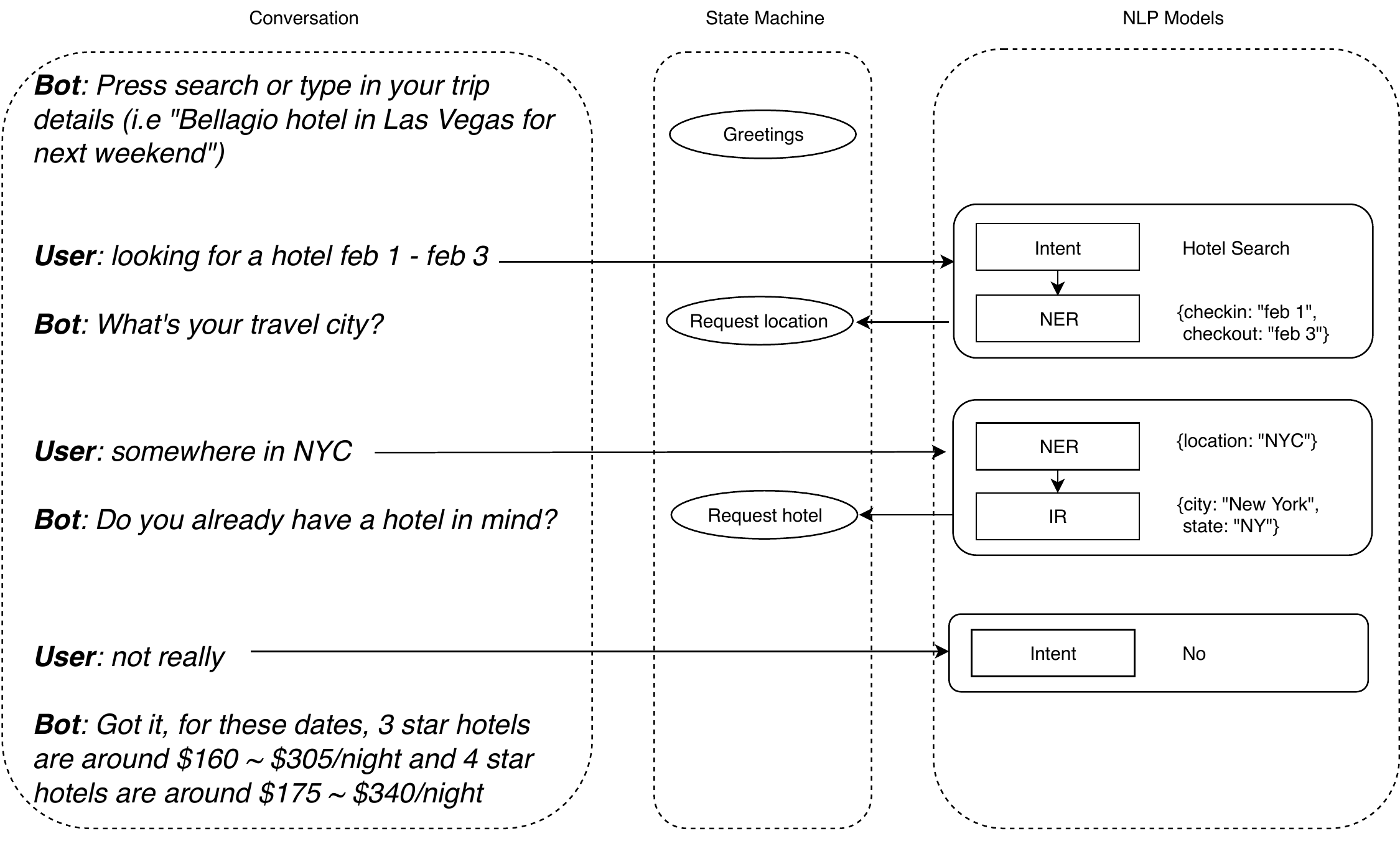}
    \caption{Example of a conversation with our bot, with corresponding state transitions and model logic. First, the user message is processed by the intent model, which classifies the message into one of several intents (described in Table \ref{table:intent-classes}). Depending on the intent and current conversation state, other models (NER and IR) may need to be invoked. Then, a response is generated based on output of the models, and the conversation transitions to a different state.}
    \label{fig:dialog-flowchart}
\end{figure*}

Our dialog system can be described as a frame-based slot-filling system, controlled by a finite-state automaton. At each stage, the bot prompts the user to fill the next slot, but supports filling a different slot, revising a previously filled slot, or filling multiple slots at once. We use machine learning to assist with this, extracting the relevant information from natural language text (Section \ref{section:models}). Additionally, the system allows universal commands that can be said at any point in the conversation, such as requesting a human agent or ending the conversation.

Figure \ref{fig:state-machine} shows part of the state machine, invoked when a user starts a new hotel search. Figure \ref{fig:dialog-flowchart} shows a typical conversation between a user and the bot, annotated with the corresponding state transitions and calls to our machine learning models.

\subsection{Data labelling}

We collect labelled training data from two sources. First, data for the intent model is extracted from conversations between users and customer support agents. To save time, the model suggests a pre-written response to the user, which the agent either accepts by clicking a button, or composes a response from scratch. This action is logged, and after being checked by a professional annotator, is added to our training data.

Second, we employ professional annotators to create training data for each of our models, using a custom-built interface. A pool of relevant messages is selected from past user conversations; each message is annotated once and checked again by a different annotator to minimize errors. We use the PyBossa\footnote{\url{https://pybossa.com}} framework to manage the annotation processes.

\section{Models}
\label{section:models}

Our conversational AI uses machine learning for three separate, cascading tasks: intent classification, named entity recognition (NER), and information retrieval (IR). That is, the intent model is run on all messages, NER is run on only a subset of messages, and IR is run on a further subset of those. In this section, we give an overview of each task's model and evaluation metrics.

\subsection{Intent model}

\begin{table}[t]
    \centering
    \caption{Some intent classes predicted by our model.}
    \begin{tabular}{|l|l|}
        \hline
         {\bf Intent} & {\bf Description} \\ \hline
         {\em thanks} & User thanks the bot \\ \hline
         {\em cancel} & Request to cancel booking \\ \hline
         {\em stop} & Stop sending messages \\ \hline
         {\em search} & Hotel search query \\ \hline
         \multicolumn{2}{|c|}{$\cdots$} \\ \hline
         {\em unknown} & Any other message \\ \hline
    \end{tabular}
    \label{table:intent-classes}
\end{table}

The intent model processes each incoming user message and classifies it as one of several intents. The most common intents are {\em thanks}, {\em cancel}, {\em stop}, {\em search}, and {\em unknown} (described in Table \ref{table:intent-classes}); these intents were chosen for automation based on volume, ease of classification, and business impact. The result of the intent model is used to determine the bot's response, what further processing is necessary (in the case of {\em search} intent), and whether to direct the conversation to a human agent (in the case of {\em unknown} intent).

We use a two-stage model; the first stage is a set of keyword-matching rules that cover some unambiguous words. The second stage is a neural classification model. We use ELMo \cite{elmo} to generate a sequence of 1024-dimensional embeddings from the text message; these embeddings are then processed with a bi-LSTM with 100-dimensional hidden layer. The hidden states produced by the bi-LSTM are then fed into a feedforward neural network, followed by a final softmax to generate a distribution over all possible output classes. If the confidence of the best prediction is below a threshold, then the message is classified as {\em unknown}. The preprocessing and training is implemented using AllenNLP \cite{allennlp}.

We evaluate our methods using per-category precision, recall, and F1 scores. These are more informative metrics than accuracy because of the class imbalance, and also because some intent classes are easier to classify than others. In particular, it is especially important to accurately classify the {\em search} intent, because more downstream models depend on this output.

\subsection{Named entity recognition}

\begin{table}
    \centering
    \caption{Results of NER Model}
    \begin{tabular}{|l|l|l|l|l|}
        \hline
         {\bf Entity Type} & {\bf Precision} & {\bf Recall} & {\bf F1}  \\ \hline
         Hotel & 0.84 & 0.53 & 0.65\\ \hline
         Location & 0.85 & 0.89 & 0.87 \\ \hline \hline
         Hotel + Location & {\bf 0.94} & {\bf 0.99} & {\bf 0.96} \\ \hline
    \end{tabular}
    \label{table:ner-results}
\end{table}

For queries identified as {\em search} intent, we perform named entity recognition (NER) to extract spans from the query representing names of hotels and cities. Recently, neural architectures have shown to be successful for NER \cite{ner-lample, ner-peters}. Typically, they are trained on the CoNLL-2003 Shared Task \cite{conll} which features four entity types (persons, organizations, locations, and miscellaneous).

Our NER model instead identifies hotel and location names, for example:

\begin{itemize}
    \item {\em ``double room in \underline{the cosmopolitan}, \underline{las vegas} for Aug 11-16''},
    \item {\em ``looking for a resort in \underline{Playa del carmen} near the beach''}.
\end{itemize}

We use SpaCy\footnote{\url{https://spacy.io}} to train custom NER models. The model initialized with SpaCy's English NER model, then fine-tuned using our data, consisting of 21K messages labelled with hotel and location entities. Our first model treats hotels and locations as separate entities, while our second model merges them and considers both hotels and locations as a single combined entity type. All models are evaluated by their precision, recall, and F1 scores for each entity type. The results are shown in Table \ref{table:ner-results}.

The combined NER model achieves the best accuracy, significantly better than the model with separate entity types. This is expected, since it only needs to identify entities as either hotel or location, without needing to distinguish them. The model is ineffective at differentiating between hotel and location names, likely because this is not always possible using syntactic properties alone; sometimes, world knowledge is required that is not available to the model.

\subsection{Information retrieval}

\begin{figure}
    \centering
    \includegraphics[width=0.9\linewidth]{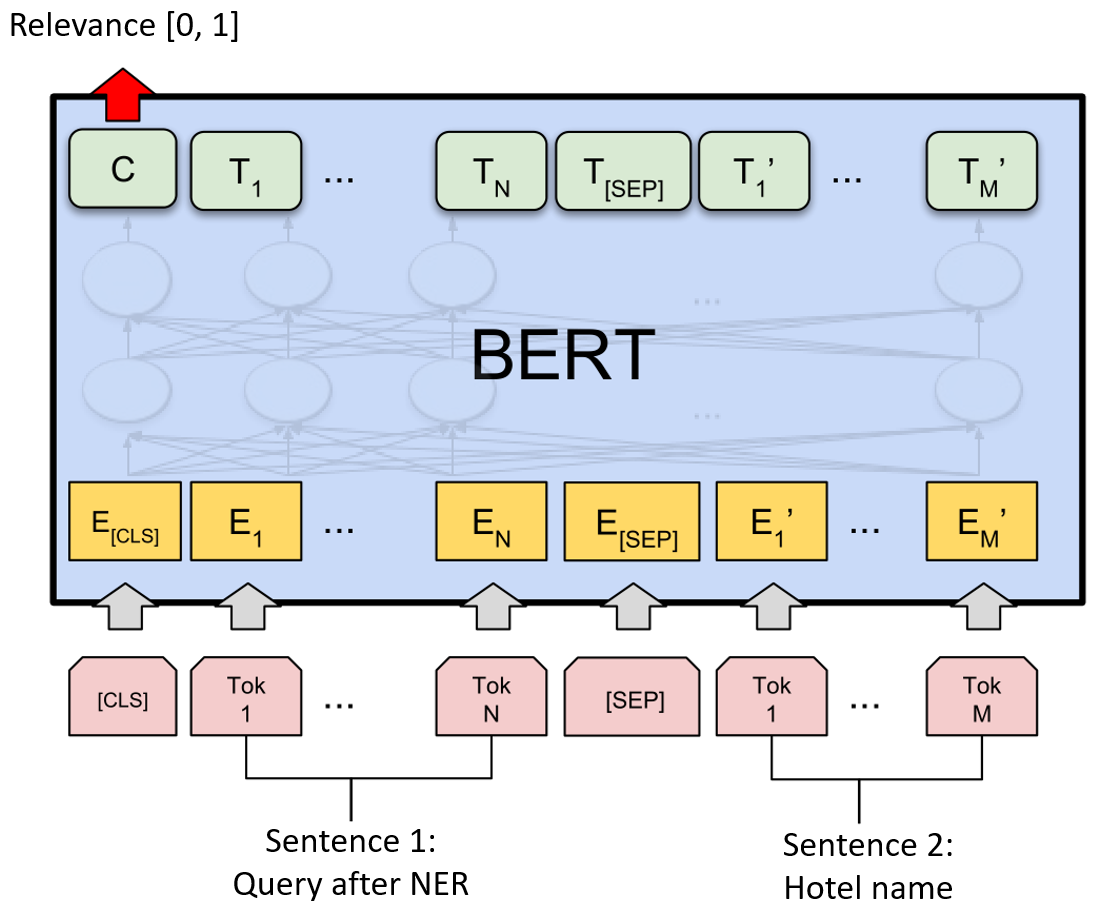}
    \caption{BERT model for IR. The inputs are tokens for the user query (after NER) and the official hotel name, separated by a \texttt{[SEP]} token. The model learns to predict a relevance score between 0 and 1 (i.e., the pointwise approach to the learning-to-rank problem). Figure adapted from \cite{bert}.}
    \label{fig:ir-diagram}
\end{figure}

The information retrieval (IR) system takes a user search query and matches it with the best location or hotel entry in our database. It is invoked when the intent model detects a {\em search} intent, and the NER model recognizes a hotel or location named entity. This is a non-trivial problem because the official name of a hotel often differs significantly from what a user typically searches. For example, a user looking for the hotel {\em ``Hyatt Regency Atlanta Downtown''} might search for {\em ``hyatt hotel atlanta''}.

We first apply NER to extract the relevant parts of the query. Then, we use ElasticSearch\footnote{\url{https://www.elastic.co}} to quickly retrieve a list of potentially relevant matches from our large database of cities and hotels, using tf-idf weighted n-gram matching. Finally, we train a neural network to rank the ElasticSearch results for relevancy, given the user query and the official hotel name.

\begin{table}
    \centering
    \caption{Results of IR Models}
    \begin{tabular}{|l|c|c|}
        \hline
         {\bf Model} & {\bf Top-1 Recall} & {\bf Top-3 Recall}  \\ \hline
         Unigram matching baseline & 0.473 & -- \\ \hline
         Averaged GloVe + feedforward & 0.680 & 0.869 \\ \hline
         BERT + fine-tuning & \bf{0.895} & \bf{0.961} \\ \hline
    \end{tabular}
    \label{table:ir-results}
\end{table}

Deep learning has been applied to short text ranking, for example, using LSTMs \cite{lstm-ranking}, or CNN-based architectures \cite{sm-cnn, mp-cnn}. We experiment with several neural architectures, which take in the user query as one input and the hotel or city name as the second input. The model is trained to classify the match as relevant or irrelevant to the query. We compare the following models:

\begin{enumerate}
    \item {\bf Averaged GloVe + feedforward}: We use 100-dimensional, trainable GloVe embeddings \cite{glove} trained on Common Crawl, and produce sentence embeddings for each of the two inputs by averaging across all tokens. The sentence embeddings are then given to a feedforward neural network to predict the label.
    \item {\bf BERT + fine-tuning}: We follow the procedure for BERT sentence pair classification. That is, we feed the query as sentence A and the hotel name as sentence B into BERT, separated by a \texttt{[SEP]} token, then take the output corresponding to the \texttt{[CLS]} token into a final linear layer to predict the label. We initialize the weights with the pretrained checkpoint and fine-tune all layers for 3 epochs (Figure \ref{fig:ir-diagram}).
\end{enumerate}

The models are trained on 9K search messages, with up to 10 results from ElasticSearch and annotations for which results are valid matches. Each training row is expanded into multiple message-result pairs, which are fed as instances to the network. For the BERT model, we use the uncased BERT-base, which requires significantly less memory than BERT-large. All models are trained end-to-end and implemented using AllenNLP \cite{allennlp}.

For evaluation, the model predicts a relevance score for each entry returned by ElasticSearch, which gives a ranking of the results. Then, we evaluate the top-1 and top-3 recall: the proportion of queries for which a correct result appears as the top-scoring match, or among the top three scoring matches, respectively. The majority of our dataset has exactly one correct match. We use these metrics because depending on the confidence score, the chatbot either sends the top match directly, or sends a set of three potential matches and asks the user to disambiguate.

We also implement a rule-based unigram matching baseline, which takes the entry with highest unigram overlap with the query string to be the top match. This model only returns the top match, so only top-1 recall is evaluated, and top-3 recall is not applicable. Both neural models outperform the baseline, but by far the best performing model is BERT with fine-tuning, which retrieves the correct match for nearly 90\% of queries (Table \ref{table:ir-results}).

\subsection{External validation}

Each of our three models is evaluated by internal cross-validation using the metrics described above; however, the conversational AI system as a whole is validated using external metrics: agent handoff rate and booking completion rate. The agent handoff rate is the proportion of conversations that involve a customer support agent; the booking completion rate is the proportion of conversations that lead to a completed hotel booking. Both are updated on a daily basis.

External metrics serve as a proxy for our NLP system's performance, since users are more likely to request an agent and less likely to complete their booking when the bot fails. Thus, an improvement in these metrics after a model deployment validates that the model functions as intended in the real world. However, both metrics are noisy and are affected by factors unrelated to NLP, such as seasonality and changes in the hotel supply chain.






\section{Conclusion}

In this paper, we give an overview of our conversational AI and NLP system for hotel bookings, which is currently deployed in the real world. We describe the various machine learning models that we employ, and the unique opportunities of developing an e-commerce chatbot in the travel industry. Currently, we are building models to handle new types of queries (e.g., a hotel question-answering system), and using multi-task learning to combine our separate models. Another ongoing challenge is improving the efficiency of our models in production: since deep language models are memory-intensive, it is important to share memory across different models. We leave the detailed analysis of these systems to future work.

Our success demonstrates that our chatbot is a viable alternative to traditional mobile and web applications for commerce. Indeed, we believe that innovations in task-oriented chatbot technology will have tremendous potential to improve consumer experience and drive business growth in new and unexplored channels.

\section{Acknowledgment}

We thank Frank Rudzicz for his helpful suggestions to drafts of this paper. We also thank the engineers at SnapTravel for building our chatbot: the conversational AI is just one of the many components.

\bibliographystyle{./bibliography/IEEEtran}
\bibliography{./acl2019.bib}

\end{document}